\documentclass[lettersize,journal]{IEEEtran}

\usepackage{array}
\usepackage[caption=false,font=normalsize,labelfont=sf,textfont=sf]{subfig}
\usepackage{textcomp}
\usepackage{stfloats}
\usepackage{url}
\usepackage{verbatim}
\usepackage{graphicx}
\usepackage{cite}
\usepackage{soul,color}

\usepackage{booktabs}
\usepackage{multirow}
\usepackage{rotating}
\usepackage{placeins}

\usepackage{hyperref} 
\usepackage{url}
\usepackage{latexsym}
\usepackage{amsmath,amsfonts}
\usepackage{algorithm}
\usepackage{algpseudocode}
\usepackage{xcolor}
\newcommand\qh[1]{\textcolor{black}{#1}}
\newcommand\qht[1]{\textcolor{black}{#1}}

\newcommand\cx[1]{\textcolor{black}{#1}}
\newcommand\cxt[1]{\textcolor{black}{#1}}
\newcommand\cxrev[1]{\textcolor{black}{#1}}
\newcommand\cxreva[1]{\textcolor{black}{#1}}
\newcommand\cxrevb[1]{\textcolor{black}{#1}}
\newcommand\cxrevc[1]{\textcolor{black}{#1}}
\begin{document}

\title{From Dense to Sparse: Event Response for Enhanced  Residential Load Forecasting}

\author{~\IEEEmembership{~IEEE}
}

\markboth{IEEE Transactions on Instrumentation and Measurement,~Accepted for publication, November~2024}%
\author{
}

\author{Xin Cao, Qinghua Tao, Yingjie Zhou$^*$, \textit{Member, IEEE,} Lu Zhang, Le Zhang, \textit{Member, IEEE,}\\ Dongjin Song, \textit{Member, IEEE,} Dapeng Oliver Wu, \textit{Fellow, IEEE,} Ce Zhu, \textit{Fellow, IEEE}
\thanks{Xin Cao (xincao@stu.scu.edu.cn) and Yingjie Zhou (yjzhou@scu.edu.cn) are 
with College of Computer Science, Sichuan University, China. Qinghua Tao is with ESAT, KU Leuven,  Belgium. Lu Zhang is with School of Cybersecurity (Xin Gu Industrial College), Chengdu University of Information Technology, China. Le Zhang and Ce Zhu are with School of Information and Communication Engineering, University of Electronic Science and Technology of China. Dongjin Song is with Department of Computer Science and Engineering, University of Connecticut, USA. Dapeng Oliver Wu is with Department of Computer Science, City University of Hong
Kong, China.  ($^*$Corresponding author: Yingjie Zhou.)}}

\maketitle
\begin{abstract}

Residential load forecasting (RLF) is crucial for resource scheduling in power systems. {
Most existing methods utilize all given \cx{load records (\textit{dense data})} to indiscriminately extract the dependencies 
between historical and future time series.
However, there exist important regular patterns residing in the event-related associations among different appliances (\textit{sparse knowledge}),  which have yet been ignored.}
In this paper, we propose an Event-Response Knowledge Guided approach (ERKG) for RLF by incorporating the estimation of electricity usage events 
for different appliances, \qh{mining 
\cxrev{event-related} sparse knowledge from the load series. With  ERKG, the event-response estimation \cxrev{enables
portraying the}} electricity consumption behaviors of residents, \qh{revealing regular 
variations} in appliance operational states.
%
%
\qh{To be specific,}
ERKG consists of 
knowledge extraction and guidance:  \qh{\emph{i)} a forecasting model is designed for the  electricity usage events by estimating appliance operational states,  aiming to extract the  event-related sparse knowledge; \emph{ii)}  a novel knowledge-guided mechanism is established by fusing  such state estimates of the appliance events into the RLF model, which can give particular focuses on 
the patterns of users' electricity consumption behaviors.} 
 Notably, ERKG can \qh{flexibly} serve as a plug-in 
 \qh{module} to boost the capability of existing forecasting models 
 \qh{by leveraging  event \cxt{response}.}
\qh{In numerical experiments, extensive comparisons and ablation studies have verified the effectiveness of our ERKG, e.g., over 8\% MAE can be reduced on the tested
state-of-the-art forecasting models.} The source code will be available at \url{https://github.com/ergoucao/ERKG}.
\end{abstract}


\section{Introduction}
\IEEEPARstart{R}{esidential} Load Forecasting (RLF) aims to predict future electricity usages for individual consumers, reflecting their anticipated household power demands and the corresponding electricity consumption behaviors.
In the power systems, different participants can be benefited from the forecast analysis of electricity \cxrevc{usage}
. 
%
For the system operator, RLF 
can provide an aggregated future power demand of residents within a certain area. Based on the current electricity resource situation and power demands, the operator formulates corresponding power dispatch or demand response strategies. For example, by adopting varied Time-of-Use pricing schemes, consumers 
 can dynamically adjust their electricity usage to achieve peak shaving and valley filling for power grid \cite{demand_benefit_system_operator_chrysopoulos2018customized}. 
For the consumer, 
RLF enables better 
residential scheduling to cope with Real-Time Pricing or to 
determine the energy storage in 
advance \cite{demand_benefit_system_consumer_morstyn2018using}. 
Accurate 
RLF facilitates an optimal allocation of power resources for boosting the  efficiency and reliability \qh{of the power systems, and thus}
has been becoming a popular research field within both the industrial and academic communities.
\cxrevc{Smart} meters deployments and Non-Invasive Load Monitoring  
advancements have provided accesses to extensive fine-grained data \cite{wang2018review, appliance_analysis_wang2021bottom}, laying the groundwork for accurate analysis of residential user behaviors. 
On such bases, many studies have been proposed to 
\qh{explore} the inherent dependencies in massive historical data \cite{kong2017LSTM, kong2017BehaviorLearning, lin2021spatial, SPatialandTemporalRLF,razghandi2020residentialAPP-lev,zhou2022appliance,SPatialandTemporalRLF}, \cxrevc{e.g.}, series historical seasonality and growth patterns.
Most existing RLF approaches can generally be categorized into two groups based on 
the forecasting levels, i.e., household level and appliance level. 
Specially, household-level methods aim at forecasting the total electricity usage for the entire  \qh{household.}
\qh{In \cite{kong2017LSTM, kong2017BehaviorLearning}, \cxrevc{the} RLF models 
address to capture the temporal dependencies from historical load series, while} 
 \cite{lin2021spatial, SPatialandTemporalRLF} utilize spatial correlations among multiple households to enhance RLF. 
Appliance-level methods focus on predicting the energy usage of electrical devices within a household, 
\qh{exploiting} the relationship between  electricity consumption habit of resident and appliance electricity usage.
RALF-LSTM \cite{razghandi2020residentialAPP-lev} proposes a recurrent deep neural network that considers the energy-saving behavior of individual users for estimating future appliance energy usage.
To improve the model capability, ALSTLF-RNN \cite{zhou2022appliance} leverages the similarity in energy consumption patterns among appliances across different households. 
In \cite{kong2017BehaviorLearning}, Kong et al. claim the mutual relationship between the  appliance-level and  household-level, and the relationship contributes to RLF.
Forecasting at both levels explicitly facilitates learning this mutual relationship. 
In fact, RLF at two levels constitutes a Multivariate Time Series Forecasting (MTSF) problem. This involves both identifying intra-variable historical patterns and modeling inter-variable relationships to reflect consumer electricity usage. However, 
the electricity usage patterns at the appliance-level and house-level are typically different, and their relationship is also dynamic \cite{rangapuram2021end}.

\begin{figure}[!t]
\centering
\includegraphics[width=3.5in]{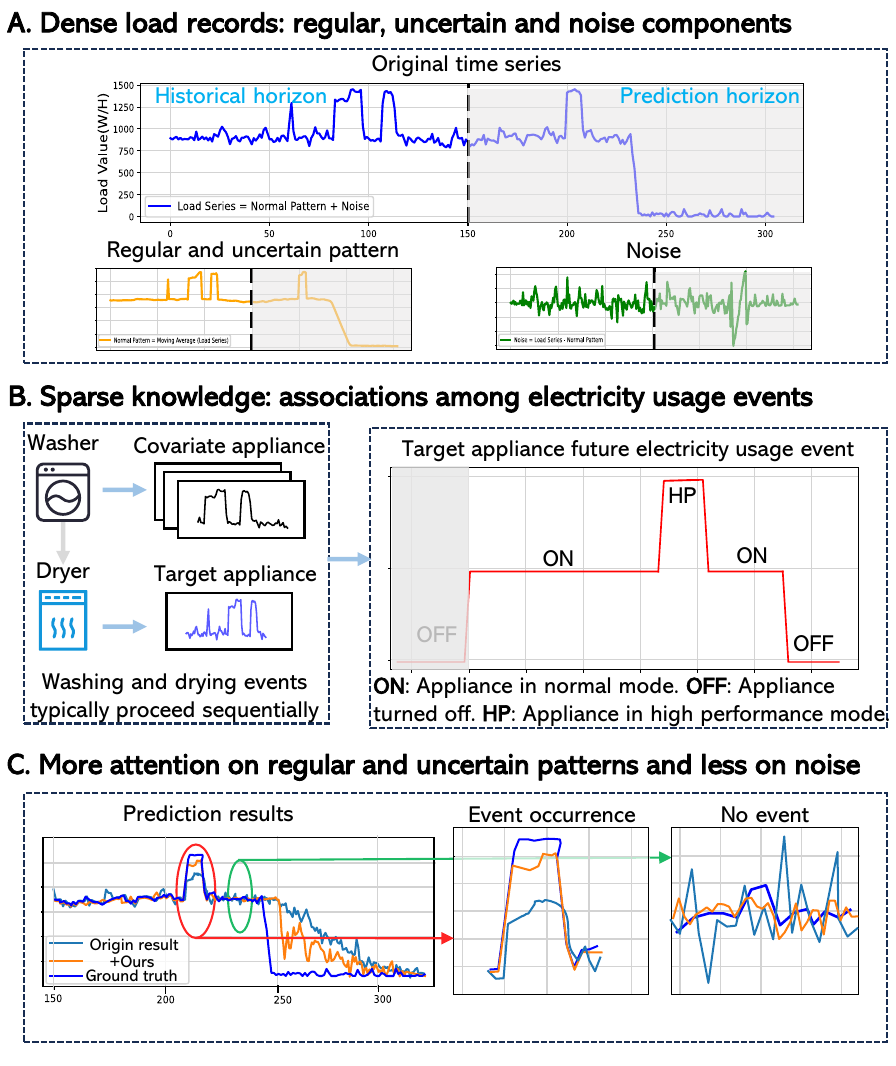}
\label{fig1}
\caption{The illustration of the proposed approach ERKG: \textbf{Learning \cxrev{event-related sparse knowledge} from dense data}, enabling model focus on the electricity consumption behaviors in reality, i.e., regular and uncertain patterns. Specifically, {(A)}: The original dense load series consist of regular and uncertain patterns, as well as noise, which impact predictive performances. {(B)}: By learning the intrinsic relationship between future electricity usage events and historical load records, ERKG extracts sparse knowledge, i.e., event-related sparse knowledge, \cxrev{which is represented by class probabilities of appliance operational states.} {(C)}: The use of sparse knowledge guides the prediction model to focus on regular and uncertain patterns, reducing noise fitting. \cxrev{In practice, we utilize appliance operational states class probabilities as weights to regularize the training loss of the prediction model.}}
\label{fig_1}
\end{figure}

Recent 
\qh{progresses} in MTSF 
\qh{have}
yielded advanced models skilled in managing complex intra-variable and inter-variable patterns, notably falling into Transformer-based and MLP-based categories.
Transformer-based models capture long-term dependency with self-attention, using positional encoding techniques to preserve temporal \cxrevc{information}\cite{zhou2021informer, Wu2021autoformer, woo2022etsformer, zhang2022crossformer}.
In particular,
Autoformer\cite{Wu2021autoformer}  proposes an Auto-Correlation mechanism based on the series periodicity, which utilizes dependencies discovery at the sub-series level to learn complex temporal patterns.
ETSformer \cite{woo2022etsformer} models series patterns by decomposing the series into interpretable components such as level, growth, and seasonality. 
Crossformer\cite{zhang2022crossformer} employs a dimension-segment embedding to comprehensively learn relationships between series variables.
On the other hand,  MLP-based models retain complete temporal information with default linear architecture.
For instance, LTSF-Linear\cite{zeng2023transformers} can be comparable to most of the previously mentioned Transformer-based models. 
This work has empirically validated that the linear model effectively retains temporal information.  
Meanwhile, IBM and Google have introduced MLP-Mixer models \cite{chen2023tsmixer, ekambaram2023tsmixer}, further improving the performance of MLP-based models. \cx{The aforementioned work indiscriminately learns all inherent patterns in given time series, assuming that the model can automatically utilize these inherent patterns for prediction.  
}


{However, residential load series blend normal patters with corruption patterns, as shown in  Fig. \ref{fig1}. A: $i$) the regular and uncertain patterns (normal patterns), which represent electricity consumption behaviors; and $ii$) the noise or anomalous disturbances (corruption patterns), potentially caused by instability and aging of appliances and measurement devices; it can clearly observed that compared to corruption patterns, the variations in normal patterns are sparse.  
Due to the patterns bias, these indiscriminate learning approaches, i.e., directly modeling the series value-level relationship between historical and future time series, often result in under-exploration of normal patterns and  overfitting of corruption patterns, leading to insufficient performance in RLF. }


We consider focusing on exploring the normal patterns based on event-related sparse knowledge, which represents the event relationships among different appliances. As shown in Fig. \ref{fig1}. B, it illustrates a novel pattern learning paradigm in RLF, which utilizes historical load records to estimate future events. Specifically, due to the cleaning habits of users, washing and drying events occur sequentially (event-related sparse knowledge). Based on this associations, when the washer (the covariate appliance) is consuming electricity and the dryer (the target appliance) is still not in operation, we can estimate the future electricity usage event of dryer, i.e., transitioning from state ``OFF'' $\Rightarrow$ ``ON'' $\Rightarrow$ ... $\Rightarrow$ ``OFF''.  This paradigm can effectively extract event-related sparse  knowledge and concentrate on the normal patterns in load series.

In this work, we propose ERKG, a Event-Response Knowledge Guided approach to enhance RLF models in performing both household level and application level load forecasting. 
To be specific, ERKG consists of knowledge extraction and guidance: \emph{\textbf{i})} for extraction, this paper propose an event estimation paradigm, it is a multivariate state predictor that forecasts the future \cxrev{appliance operational states} based on historical load series, and specifically models the state changes of appliances to learn event-related sparse knowledge; \emph{\textbf{ii})} for guidance, this paper develops a knowledge guide mechanism based on event response to fuse event-related knowledge into the RLF model. Notably, ERKG can flexibly serve as a plug-in module to enhance the concentration of RLF model training on key areas, i.e., event occurrence in Fig. \ref{fig1}. C, without changing the model’s structure and inference process.
Finally, experiments on three datasets with three state-of-the-art prediction models demonstrate a 8\% improvement in Mean Absolute Error (MAE), affirming the effectiveness of  ERKG. The main contributions of this work are summarized as follows.

\begin{enumerate}
    \item  We propose an event response knowledge-guided approach to address the challenge of 
    \qh{exploiting}  event-related sparse knowledge from
    \qht{the given load records (dense data)} for  \qh{RLF}.

\item 
\qh{For} event-related sparse knowledge extraction, an forecasting model is designed for 
electricity usage events by estimating 
the operational states of \qht{appliances}, 
\qht{obtaining} event-related associations 
from historical load series. 

\item With such event-related sparse knowledge, a novel knowledge-guided mechanism \qht{is constructed by leveraging} 
event response to promote the learning of regular and uncertain patterns that are beneficial for forecasting.

\item  Extensive numerical experiments have been conducted to evaluate the proposed ERKG on widely used residential datasets, 
\qht{showing}
that ERKG notably enhances the  performance of existing state-of-the-art forecasting models as a plug-in module.
\end{enumerate}

    The remainder of this work is organized as follows. Section $\textup{II}$ presents the related works. The problem definition is elaborated in Section $\textup{III}$.  Section $\textup{IV}$ introduces the proposed method. Section $\textup{V}$ gives experimental results. We conclude our work with outlooks in Section $\textup{VI}$.

{
}


{
}

\section{RELATED WORK}\label{sec:related work}

\subsection{Residential Load Forecasting}
The residential load is a 
\qh{typical type of} time series, \qh{for which many forecasting methods have been proposed, involving traditional statistical and machine learning techniques, where deep learning methods are currently of popularity.}
Traditional statistical methods can learn linear or simple nonlinear relations in data through linear combinations of historical data, e.g., the AutoRegressive Integrated Moving Average (ARIMA) method.
Specifically, CSTLF-ARIMA \cite{statis_ARIMAforRLF} integrates the ARIMA model and the transfer function to improve model prediction accuracy, where the relations between weather and load are taken into account.
\cxrevb{RLF-IGA \cite{statis_GaussianProcesforRLF} employs an integrated Gaussian process that leverages associations from both target and related customers for reliable hourly residential load prediction.}
Through specialized feature engineering techniques, machine learning methods can handle more complicated nonlinearity in data, e.g., Support Vector Regression (SVR) machines.
MFRLF-SVR \cite{jain2014forecasting} 
uses SVR to investigate the impact of temporal granularity on model accuracy, suggesting that the optimal granularity is hourly.
\cxrevc{HFSM} \cite{HFSM} proposes a  
holistic feature selection method, boosting ANN-based forecasting model. 
In particular, recent advancements
show that deep learning methods are highly effective in capturing complex patterns residing in time series. \cx{Particularly}, Long Short-Term Memory \cite{kong2017LSTM} leverages
recurrent network structures to retain complex temporal patterns.  
Meanwhile, many methods propose
model spatial relationships among residential load series, e.g., the relations among residential users \cite{lin2021spatial, SPatialandTemporalRLF} and the associations between appliances \cite{razghandi2020residentialAPP-lev,zhou2022appliance}.
It is worth noting that the electrical appliance load plays a positive role in household load forecasting \cite{kong2017BehaviorLearning}.
However, only a few studies have analyzed the electrical characteristics of appliances.
For instance, 
MTL-GRU \cite{appliance_analysis_wang2021bottom} categorizes appliances into continuous and intermittent load appliances and employs different methods for processing them. 
\cxrev{\cxrevc{Meanwhile}, Welikala et al. \cite{welikala2017incorporating} incorporate appliance usage patterns (AUPs) for NILM and load forecasting. Their approach is based on pre-observed AUPs to obtain the prior probabilities of appliances being activated (``ON''), resulting in more accurate NILM and load forecasting results.}

\cxrev{Our ERKG leverages an electricity usage event forecasting model that learns not only complex intra-variable patterns, e.g., changes in appliance usage patterns over different times, but also inter-variable patterns, e.g., relationships between different appliances. Specifically, different from these work \cite{appliance_analysis_wang2021bottom,appliance_analysis_qian2020variable} that focuses on the characteristics of individual appliances or the combination of activity/inactivity of appliances \cite{welikala2017incorporating}, our work considers the event-related associations among appliances with multiple different appliance operating states to enhance state-of-the-art MTSF models for RLF.}

\subsection{Knowledge Guided Deep Learning}

Deep learning methods have been  proven powerful in various areas, such as computer vision and recommending systems. However, when 
special scenario events occur, such as traffic flow under inclement weather and accidents, relying solely on 
the given raw data can make it difficult to discover important factors and learn complex data relationships.
Thus, the resulting performances can still be limited, depite of the great model capacity of deep learning methods.
Some existing work \cite{invasive_knowledge_qi2021known, invasive_knowledge_yin2019domain,invasive_knowledge_peng2019knowledge,liu2022fedtadbench, RobustTrafficPrediction} propose 
to utilize prior knowledge to guide the learning process.
Yin et al. \cite{invasive_knowledge_yin2019domain} use accumulated medical knowledge from large public datasets to compensate for data in clinical car, while Peng et al. \cite{invasive_knowledge_peng2019knowledge} 
utilize context information from events, e.g., accidents, and environmental factors,
 e.g., weather,
 to construct knowledge graph that guide the learning of traffic flow predictors.
These methods can be categorized into two types, depending on the utilization of external knowledge or internal model knowledge.

For the methods based on external knowledge, the performances are enhanced by learning more specific domain data. 
Kerl \cite{invasive_knowledge_wang2020kerl} is a knowledge-guided reinforcement learning model that integrates knowledge graph information 
for sequential recommendation, 
\cxrev{facilitating the handling of} sparse and complex user-item interaction data. 
\cxrevb{Peng et al. \cite{invasive_knowledge_peng2019knowledge} demonstrate using knowledge and situational awareness, i.e., weather and traffic data, to improve model performance under conditions such as severe traffic accidents.}
\cxrevb{Yin et al. \cite{invasive_knowledge_yin2019domain} propose a domain knowledge-guided network that utilizes the causal relationships among clinical events to improve modeling accuracy.}
\cxrevb{These} approaches require custom embedding and design of the original model based on specific knowledge, limiting their extensibility.

For the methods based on internal model knowledge, the learning 
relies on utilizing knowledge 
from auxiliary models. In particular, to improve  the efficiency of knowledge-guided teacher-student (auxiliary-target) learning, Hinton et al. \cite{non_invasive_knowledge_hinton2015distilling} introduce response-based knowledge, i.e., soft targets, for image classification, which preserves the probabilistic relationships among different categories.
Considering task-specific knowledge in multi-task recommendation, Yang et al. \cite{non_invasive_knowledge_yang2022cross} propose a cross-task learning framework that uses response-based knowledge contained in auxiliary tasks. 
\cx{Response-based knowledg}e guide methods are simple yet effective, which directly utilize logits regularization to make the student (target) model mimic the neural response of the  teacher (auxiliary) model. The logits regularization loss does not affect the structure and inference of the target model. The proposed ERKG is an event response-based method that can serve as a plug-in module. Unlike the above response-based methods, we use event-related sparse knowledge to construct conditional targets to guide the target model in identifying and focusing on key patterns within the data.

\section{Problem Definition}
\begin{figure}
    \centering
    \includegraphics[width=3.5in]{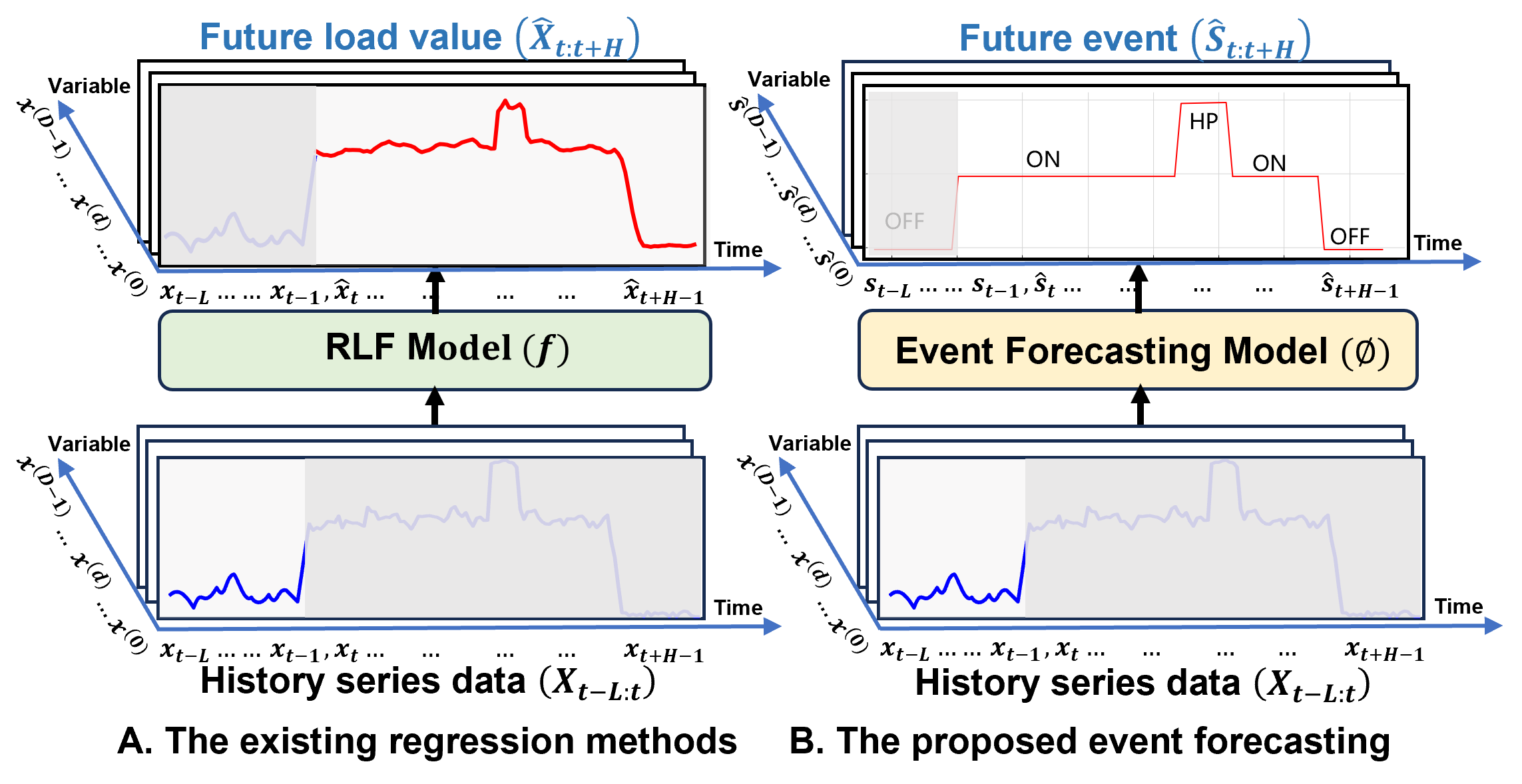}
     \vspace{-10pt}
    \caption{Two learning paradigms. (A): 
    \cxrevb{Existing methods directly predict future loads using historical load values.}
    (B): \cxrevb{Our event forecasting method estimates appliance operational states using historical load values and models the state changes of appliances to learn event-related sparse knowledge.}}
    \label{fig:fig_paradigm}
\end{figure}
Our approach involves conducting RLF at both household level and application level simultaneously.  
Let $x_{t}=[x_t^{(0)},\dots,x_t^{(D-1)}]$ denote an observation of household and appliances
load series at time step $t$, where $D$ equals  of smart meters, including measuring devices
for appliances and household rooms. As shown in Fig. \ref{fig:fig_paradigm}. A,  given a lookback window
$X_{t-L:t}=[x_{t-L},\dots,x_{t-1}]$, 
we consider the task of predicting future $H$-length time steps, $X_{t:t+H}=[x_t,\dots,x_{t+H-1}]$. 
We denote $\hat{X}_{t:t+H}$ as the point forecast of $X_{t:t+H}$. Thus, the goal is to learn
a forecasting function $f$ by minimizing some loss function 
$\mathcal{L}:\mathbb{R}^{H\times D}\times\mathbb{R}^{H\times D}\to\mathbb{R}$,  as shown in Eq. \eqref{eq:forecasting_function}:

\begin{equation}
\label{eq:forecasting_function}
\hat{X}_{t:t+H}=f(X_{t-L:t}).
\end{equation}

In previous work, load values are often treated as single entities, and modeling these values is considered as a regression task\cxrev{. 
For example, VAE-LF \cite{langevin2023efficient} 
utilizes a variational autoencoder-based regression model to describe the
local relationship between historical aggregated load values and future target
appliance values, with the target appliances including manually pre-selected
ones, i.e., Washing Machine, Dishwasher, and Fridge.
}
However, regression-based methods, along with the corresponding
loss function such as MSE, face limitations. Notably, due to the double penalty issue\cite{double-penalty}, they tend to produce conservative forecasts for future load curves\cite{xia2023day}, which may overlook event-related knowledge. To overcome this limitation, as shown in Fig. \ref{fig:fig_paradigm}. B,  we transform continuous load values into states for predicting \cxrev{appliance operational states}.  This task is inherently a multi-task forecasting problem due to the concurrent operations of multiple appliances.
\cxrev{Different from VAE-LF that employing a prediction of future load values as an estimation of operational states, our ERKG first identifies the various multi-states of different appliances through adaptive clustering and then overviews historical data to predict future states as a description of electricity usage events.}

Let $x^{(i)}=[x_0^{(i)},\dots,x_{l-1}^{(i)}]$ denote the load series for the i-th variable where $i\in{\{0,\dots,D-1\}}$, $l$ equals the origin series length, and $n_{x^{(i)}}$ represents the number of operation states for variable $x^{(i)}$, e.g., $n_{\rm{dryer}}$ equals three in Fig. \ref{fig_1}. B: ``OFF'', ``ON'' and ``HP''. We define $s_t \in\mathbb{R}^D $ as  the state for $x_t$, with each $s_{t,i}\in{\{0,\dots,n_{x^{(i)}}-1\}}$ corresponding to $x_t^{(i)}$. Likewise, given a lookback window $X_{t-L:t}$, and a set of  operation state numbers $N_{0:D}=\{ n_{x^{0}},\dots,n_{x^{D-1}}\}$, $\hat{S}_{t:t+H}$ denotes the forecast of the state $S_{t:t+H}=[s_t,\dots,s_{t+H-1}]$, obtained by minimizing the cross-entry loss function. The state forecasting function $\phi$ is defined in Eq. \eqref{eq:state_forecasting_function}:  
\begin{gather}
\label{eq:state_forecasting_function}
\hat{S}_{t:t+H}=\phi{\left(X_{{t-L}:t}\right)}.
\end{gather}

Eq. \eqref{eq:forecasting_function} and Eq. \eqref{eq:state_forecasting_function} represent two learning paradigms, as shown in Fig. \ref{fig:fig_paradigm}. In contrast to Eq. \eqref{eq:forecasting_function}, learning directly from the series value-leve (Fig. \ref{fig:fig_paradigm}. A), Eq. \eqref{eq:state_forecasting_function} is a new paradigm that models electricity consumption behaviors through electricity usage events, which facilitates RLF (Fig. \ref{fig:fig_paradigm}. B).

{
}
\begin{figure*}

    \centering
    \includegraphics[width=6in]{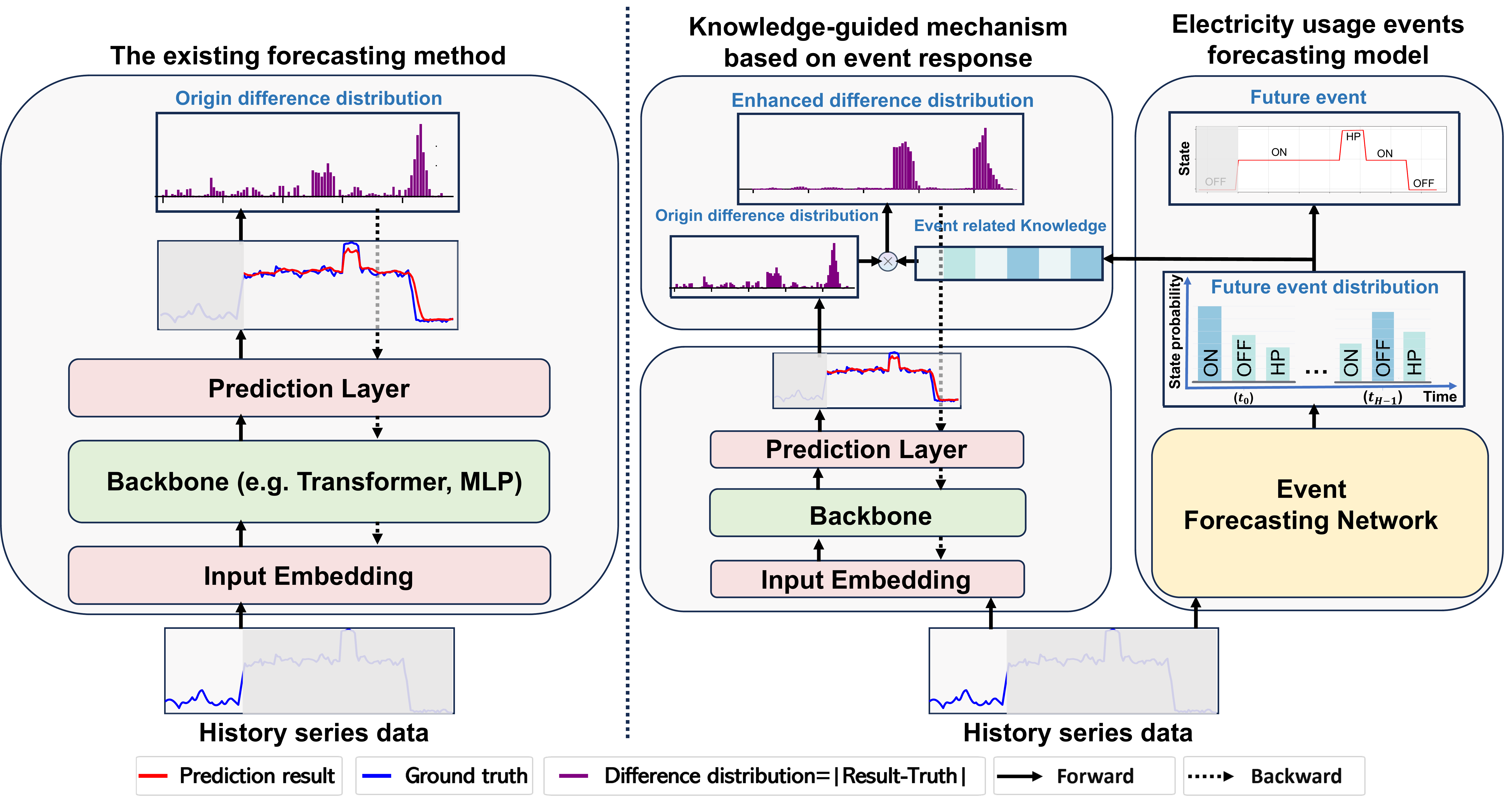}
 \vspace{-10pt}
    \caption{\qh{\emph{Left Penal}}: The 
 paradigm of existing methods, e.g., Crossformer\cite{zhang2022crossformer}, ETSformer\cite{woo2022etsformer} and TSMixer\cite{ekambaram2023tsmixer}. They directly model series value-level relationship between history and future. \qh{\emph{Right Penal}}: The overview of the proposed ERKG Approach. It can enhance the performance of 
existing methods through the estimation
of electricity usage events for different appliances }
    \label{fig2}
\end{figure*}
\section{METHODOLOGY}\label{sec:section2}

\qh{
In this section, 
a novel knowledge guided approach  for RLF introduced, namely ERKG, which 
gives particular considerations to 
event-related sparse knowledge from the given dense load records
and meanwhile avoids the overfitting to noise. 
As illustrated in Fig. \ref{fig2}, ERKG consists of  two main parts, i.e., a 
electricity usage events forecasting model and a knowledge-guided mechanism based on \cx{event response}. The former
exploits event-related sparse knowledge; the latter  utilizes such sparse knowledge to extract the key patterns residing in the data and meanwhile to reduce attention on noises. 
Notably, ERKG can be flexibly plugged into various RLF models, such as Crossformer, Etsformer, and TSMixer, as explained in \ref{enhancement procedure}. 
}

{
}
\begin{figure}
    \centering
    \includegraphics[width=3.5in]{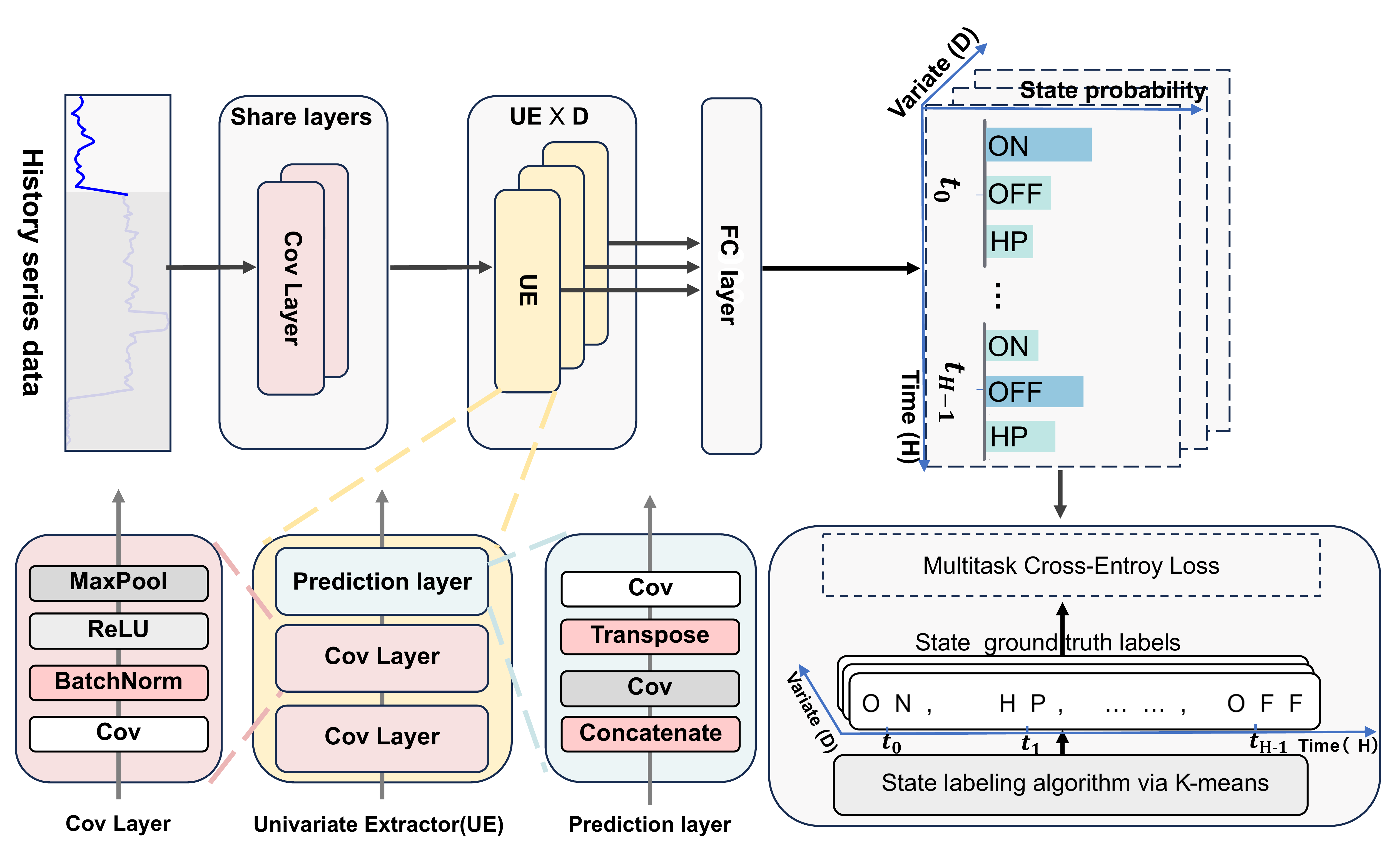}
     \vspace{-10pt}
    \caption{The proposed electricity usage events forecasting model. }
    \label{fig3}
\end{figure}
\subsection{Electricity \qh{U}sage \qh{E}vent \qh{F}orecasting \qh{M}odel}
For event-related sparse knowledge extraction, the electricity usage events forecasting model estimates appliance operational states, focusing on event-related associations in historical load series \cxrev{by training the Multivariate State Predictor (MSP) to predict the probability of future electricity usage events.} As shown in Fig. \ref{fig3}, it includes the state predictor and the state labeling algorithm: $i)$ a Multivariate State Predictor (MSP) is created to predict the future state class probabilities of series, e.g., ``ON'', ``OFF'', and ``HP'' ; $ii)$ a state labeling algorithm uses $k$-means to obtain the corresponding state ground truth labels, which facilitates the predictor in learning sparse knowledge.

\subsubsection{Multivariate State Predictor}
The MSP processes historical multivariate load series values to predict the corresponding future multi-step series state class probabilities. It processes  $D$ variates and an $L$-length load series $X_{t-L:t}=\{X_{t-L:t}^{(i)} \mid i\in \{0,1,\dots,D-1\}\}$   to produce an $H$-length load state series $\hat{S}_{t:t+H}=\{\hat{S}_{t:t+H}^{(i)}\mid i\in \{0,1,\dots,D-1\}\} $. 
The MSP consists of three components in sequence: Shared layers, $D$ Univariate Extractors (UE), and a Fully Connected (FC) layer, denoted as $\phi_{sl}$ , $\phi_{u} = \{\phi_{u^{(i)}} \mid i \in \{ 0, 1, \dotsc, D-1\}\}$,  $\phi_{fc}
$, respectively.  Specifically,  $\phi_{sl}$ extract low-level information, \qh{and $\phi_{u^{(i)}}$ is} responsible for independent state predictions with $X_{t-L:t}$, \qh{such that}: 
\begin{gather}
Z_u^{(i)} = \phi_{u_{x^{(i)}}}{(\phi_{sl}(X_{t-L:t}))}, 
\text{ for } i \in \{0, 1, \ldots, D-1\} \label{equal_ue},
\end{gather}
\qh{where $Z_u^{(i)}$ is the probability score distributions of  univariate $x^{(i)}\in \mathbb R^{H \times n_{x^{(i)}}}$, and} $n_{x^{(i)}}$  equals the number of \cxrev{appliance} operation states for variable $x^{(i)}$. 
\qh{Aligned with the feature processing of data in} \cite{CBR}, we use one-dimensional (1-$D$) convolution to construct the Prediction layer in UE.


\qh{Concerning the connections among appliance states, we apply} 
$\phi_{fc}$ to integrate
their 
\qh{correlations as in} Eq.\eqref{euqal_logit}: 
\begin{gather}
Z =  \phi_{fc}(Z_u) \label{euqal_logit},
\end{gather}
where we concatenate all $Z_u^{(i)}$ to obtain $Z_u$ $\in \mathbb{R}^{H\times \sum_{i=0}^{D-1}{n_{x^{(i)}}}}$  and pass $Z_u$  through  $\phi_{fc}$ to get $Z \in \mathbb{R}^{H\times \sum_{i=0}^{D-1}{n_{x^{(i)}}}} $. 

We then split $Z$ into $D$ group probability score distributions  $\{Z^{(i)} \mid i \in \{0,1,\dots,D-1\}\}$ to calculate the state category for each variable. In fact, the MSP concludes $\hat{S}_{t:t+H}=\{\hat{S}_{t:t+H}^{(i)} \mid i \in \{0,1,\dots, D-1\} \} $  with maximum probability score $Z^{(i)}$. 
\qh{We thereby consider} the load state of each variable at each time step as follows:  
\begin{gather}
\hat{s}_{\tau}^{(i)} = \mathop{\arg\max}\nolimits_{c}{z^{\tau,i,c}}, \text{ for } c \in \{0,1,\dotsc,n_{x^{(i)}}\}, \label{equal_argmax}
\end{gather}
where $\hat{s}_{\tau}^{(i)}$  is the load state of the $i$-th variable  at time step $\tau$, and $z^{\tau,i,c}$ is the score of the corresponding $c$-th state. \qh{L}et $\hat{S}_{t:t+H}^{(i)}=\{\hat{s}_{\tau}^{(i)} \mid \tau \in \{0,1,\dots,H-1\}\}$, and we have the $H$-length load state series  $\hat{S}_{t:t+H}=\{\hat{S}_{t:t+H}^{(i)} \mid i \in \{0,1,\dots, D-1\} \}$ for all $D$ variables $X_{t-L:t}$.

For MSP, we use a softmax function for normalization, \qh{and} then choose multitask cross-entropy as our loss function $\mathcal{L}_{\rm MSP}$, which is defined as:
\begin{gather}
p_{\tau,i,c}=\frac{e^{z_{\tau,i,c}}}{\sum_{\ell=0}^{n_{x^{(i)}}-1}{e^{z_{\tau,i,\ell}}}},\\
\mathcal{L}_{\rm MSP}=\frac{1}{H} \sum_{\tau=t}^{t+H-1}\frac{1}{D}\sum_{d=0}^{D-1}-\log({p_{\tau,i,s_{\tau}^{(i)}}}), \label{event_forecasting_model}
\end{gather}
where $p_{\tau,i,c}$ is the probability score for the 
\qh{$c$-th} category of the 
\qh{$i$-th} variable at future time step 
$\tau$, 
calculated using the softmax function, $H$ is the number of forecasting time steps, $D$ is the number of dimension variables, and $s_{\tau}^{(i)} \in \{0,1,\dots,n_{x^(i)}\}$ is the state ground truth 
label obtained by State Labeling Identification Algorithm for the 
\qh{$i$-th} variable at time step $\tau$.

\subsubsection{State Labeling Identification Algorithm}
The state identification algorithm aims to address the issue of obtaining the state class of series values by performing \cxrev{an} overall analysis of the load series.
In fact, to train the MSP effectively, we utilize $k$-means clustering\cite{TSLearning} to categorize the original series values into distinct states\cite{liu2023unsupervised}, forming the state ground truth labels $S\in{\mathbb{R}^{l\times{D}}}$.
Here, $N_{0:D}$ denotes number of state categories. Specifically, given the original load series $\mathcal{E} \in \mathbb{R}^{l\times D}$, where $l$ is the length of time step and  $D$ represents  the variable dimensionality of load series (indicative of the number of appliances and rooms). We employ sliding windows of size $w$ to generate samples and set the bounds for clustering states as $min_s$ and $max_s$, \cxrev{referring to the existing studies\cite{stateNumducange2014novel,stateNumyu2016nonintrusive} on the number of appliance operational states, we set  $min_s$ to 2 and $max_s$ to 5}. \qh{Then, the} $k$-means clustering is  applied to each variable, \qh{where} the optimal number of clusters $k_{\text{max}}$ is determined as the final state number $N_{0:D}$ based on the silhouette score\cite{rousseeuw1987silhouettes}, \qh{as  presented} in Algorithm \ref{clustering algorithm}.


\begin{algorithm}
\caption{State Labeling Identification via Time Series Clustering}
\label{clustering algorithm}
\begin{algorithmic}[1]
\State \textbf{Input:} origin load series $\mathcal{X}$; origin load time length $l$; The variables dimension of load series, $D$; windows size $w$, state minimum $min_s$, state maximum $max_s$
\State \textbf{Output:} State Profile $S$

\State $\mathcal{E} \gets \Call{ZeroMatrix}{l, w, D}$
\For{$t \in \{0, \ldots, l - w\}$}
    \State $\mathcal{E}[t] \gets \mathcal{X}[t:t + w]$
\EndFor
\For{$t \in \{l - w + 1, \ldots, l\}$}
    \State $\mathcal{E}[t] \gets \mathcal{X}[t - w:t]$
\EndFor

\State $S \gets \Call{ZeroLike}{\mathcal{X}}, N \gets \Call{ZeroMATRIX}{{D}}$
\For{$i \in \{1, \ldots, D\}$}
    \State $M_{\text{max}}, S_{\text{max}}, k_{\text{max}} \gets 0, \text{nil}, \text{nil}$
    \For{$k \in \{min_s, \ldots, max_s\}$}
        \State $s \gets \Call{TimeSeriesKMeans}{n}.\Call{Fit}{\mathcal{E}[:,:,i]}$
        \State $m \gets \Call{SilhouetteScore}{\mathcal{E}[:,:,i], s}$
        \If{$m > m_{\text{max}}$}
            \State $m_{\text{max}}, s_{\text{max}}, k_{\text{max}}\gets m, s, k$
        \EndIf
    \EndFor
    \State $S[:, i], N[i] \gets s_{\text{max}}, k_{\text{max}}$
\EndFor
\State \Call{Output}{$S$, $N$}

\end{algorithmic}
\end{algorithm}


\vspace{-10pt}



\subsection{Knowledge-Guided Mechanism Based on Event Response}
\qh{Based on event response, we propose a knowledge-guided mechanism to guide the RLF model with more attention on normal patterns of the load series and less sensitivity to the corrupted patterns, thereby enhancing the model prediction performances.}
More specifically, this mechanism employs event-related knowledge to enable the RLF model to identify and 
\qh{focus on informative event-related patterns.} \cxrev{We use the values of the multi-step series state class probabilities as weights (indicating event-related knowledge) to regularize the traditional MAE or MSE training loss, through a specifically designed Teacher-Student (T-S) learning process. 
For example,} more loss penalties on key areas, while less on areas that may contain noise. Notably, a key area can be the place where the state changes (event occurrence) , which can be observed in the comparison between the difference distribution and
enhanced difference distribution shown in Fig. \ref{fig2}.

In particular, we design a Teacher-Student (T-S) learning process to transfer event-related knowledge from a 
\qh{MSP (Teacher)
to the RLF model 
(Student).} In the T-S process,  the response-based knowledge usually refers to the neural response of the last output layer of the teacher model. The main idea is to let the \qh{S}tudent model directly mimic the final prediction of the \qh{T}eacher model. Given a vector of logits
$Z$ as the outputs of the last output layer of a deep model, the loss for response-based knowledge can be formulated as
\begin{gather}
L_{Res}(Z_t,Z_s) = \mathcal{L}_R(Z_t, Z_s),
\end{gather}
where $\mathcal{L}_R$ indicates the divergence loss of logits, and $Z_t$ and $Z_s$ are logits of \qh{Teacher and Student, respectively}. Different from the general response-based knowledge transfer in T-S process, 
\qh{we} utilize the
\qh{MSP} to guide the Student to focus on key area of learning through $Z_t$, instead of directly having the Student mimic the Teacher. Moreover, our Teacher and Student 
\qh{have} different forms outputs. Therefore, we design a response-based event-related knowledge loss as 
\begin{gather}
\mathcal{L}_{\rm Res}(Z_t,\hat{Y}, Y) = \mathcal{L}_{\rm PR}(\hat{Y}, Y | Z_t),
\end{gather}
where the $\mathcal{L}_{\rm PR}$ indicates the error between the predicted and actual values in the space specified by the $Z_t$ logit. Specifically, we denote 
$Z_t$ as $Z$, with the logits $Z$ in Eq.\eqref{euqal_logit} we define $\mathcal{L_{\rm PR}}$ as
\begin{gather}
\mathcal{L}_{\rm PR} = \frac{1}{D}\sum_{i=0}^{D-1}\frac{1}{H}\sum_{j=t}^{t+H}{{\max(Z_j^{(i)})}|\hat{Y}_j^{(i)}-Y_j^{(i)}|},\\
\mathcal{L}_{\rm MAE} = |\hat{Y}-Y|,\\
\mathcal{L} = \mathcal{L}_{\rm MAE} + \alpha \mathcal{L}_{\rm PR},\label{guide_loss}
\end{gather}
where $\hat{Y}_j^{(i)}$ and $Y_j^{(i)}$ are respectively predicted and actual values at the $j$-th time step and $i$-th dimension variable, $H$ represents the number of forecasting time
steps, and $D$ represents the number of dimension variables.
\qh{The maximization operation on $(Z_j^{(i)})$ acts as} a weight 
\qh{emphasizing the} key area in the learning process. 
$\mathcal{L}_{\rm PR}$ indicate the $Z$ of MSP specifies a weight on different time step and dimension, it encourage the model to focus on key area. $\mathcal{L}_{\rm MAE}$ is \qh{a} general regression loss, and $\mathcal{L}$ is \qh{the} final loss to train the RLF model.

\subsection{\qh{Integration as a Plug-in Module to RLF Models}}\label{enhancement procedure}
\qh{Our proposed ERKG} 
\qh{can be flexibly plugged into} most neural network methods for multi-step time series forecasting 
\qh{with regression losses, such as} Mean Absolute Error (MAE), Mean Squared Error (MSE). Firstly, as introduced in Section III.A, we utilize
the State Labeling Identification Algorithm to obtain state ground truth labels that indicate event information, and we train
the electricity usage events forecasting model end-to-end using these labels. Secondly, the RLF model training process is guided by a knowledge-guided mechanism based on event response, as detailed in Section III.B, to enhance the predictive performance. The enhancing procedure is presented in Algorithm \ref{training-algorithm}.

\begin{algorithm}
\caption{Enhancing Procedure}
\label{training-algorithm}
\begin{algorithmic}[1]
\State \textbf{Input:} Input series $X_{{t-L}:t}$ for MSP $\phi$, max\_epochs, hyperparameters $\alpha$
\State \textbf{Output:}  RLF model $f$
\State Initialize model parameters: $\phi$, $f$
\State \textbf{Stage 1: train the MSP}
\State Get state ground truth labels $S$ by algorithm \ref{clustering algorithm}
\Repeat
    \For{$t \in \{1, \ldots, T\}$}
        \State Obtain $\hat{S}_{t:t+H}$ from $\phi$ for input $X_{{t-L}:t}$
        \State Compute the MSP loss  $L_{\text{MSP}}$ (Eq. \ref{event_forecasting_model}) with $S_{t:t+H}$
        \State Update the parameters of $\phi$ using $L_{\text{MSP}}$
    \EndFor    
\Until{converge}

\State \textbf{Stage 2: Knowledge-guided RLF model training process}
\Repeat
    \For{$t \in \{1, \ldots, T\}$}
        \State Fix $\phi$ and input $X_{{t-L}:t}$ to obtain $Z=\phi(X_{{t-L}:t})$
        \State Compute the knowledge-guide loss  $ L$ using Eq. \ref{guide_loss}
        \State Update the parameters of $f$ using $L$
    \EndFor
\Until{converge}
\State \textbf{return} 
enhanced model $f$
\end{algorithmic}
\end{algorithm}

\vspace{-5pt}





\section{Experiments and Discussion\qh{s}}
\subsection{Datasets}  \label{sec:III-A}
In the experiments, we evaluate the proposed method on three publicly available and actual residential load datasets, i.e., Ampds2\cite{makonin2016electricity}, UK-Dale\cite{kelly2015uk}, and UMass Smart Home\cite{barker2012smart}. They encompass both household-level aggregated and appliance-level disaggregated power data. 
\qh{More descriptions} of the three datasets are summarized in Table \ref{table:dataset_des}. 

\paragraph{Almanac of Minutely Power Dataset \cite{makonin2016electricity}} AMPds2 collected energy consumption data of a residential house with 20 appliances and weather data in Canada from April 2012 to April 2014. It contains a total of 1,051,200 readings from two years of monitoring per meter. 

\paragraph{UK Domestic Appliance-Level 
Electricity Dataset \cite{kelly2015uk}}
UK-Dale is an open-access dataset from the UK recording domestic appliance-level electricity. 
UK-Dale contains data for five actual residential houses in the UK, 
and we use the data from household 1 for the experiment, which was recorded for 655 days, and 47 dimension measure result.

\paragraph{UMass Smart Home Dataset \cite{barker2012smart}} Multiple smart meter readings of 7 homes were collected by the UMass Smart Home project in the US from 2014 to 2016,
\qh{containing} readings from separate smart meters controlling different appliances, \qh{where}
we 
\qh{choose} houses C and D for the experiments.

\qh{Before training,}
a uniform preprocessing procedure has been performed on each dataset. Firstly,
we align the start and end times of all measurement series within the same house, including both household-level and appliance-level. We also 
utilize downsampling to standardize the sampling period of series to 
1 hour, \cxrev{meaning that each time point interval in the input model samples is 1 hour}. Then, through the process of algorithm \ref{clustering algorithm}  for state labeling identification, the state data for all series are obtained. Finally, 60\% of the data are used for training, 20\% are used for validation, and the remaining 20\% are used for testing. The data were normalized using the mean and standard deviation values computed from the training set. 

\vspace{-10pt}
\begin{table}[htbp]
  \centering
    \caption{\qh{Descriptions of the datasets}}
    \begin{tabular}{cccc}
    \toprule
    Dataset & Range & Sample Rate & Dimension \\
    \midrule
    Ampds2 & 2012/4/01-2014/4/01 & 1 minute & 23 \\
    UK-DALE & 2013/4/11-2014/12/23 & 1 minute & 47 \\
    \cxrev{UmassC} & 2016/1/01-2016/12/15 & 1 minute & 22 \\
    \cxrev{UmassD} & 2016/1/01-2016/12/31 & 1 hour    & 72 \\
    \bottomrule
    \end{tabular}%
  \label{table:dataset_des}%
\end{table}%

\subsection{Performance Metrics} \label{sec:III-B}
We evaluate all RLF methods in the experiments with two 
performance metrics, i.e., 
the Mean Absolute Error (MAE) and the symmetric Mean Absolute Percentage Error (MAPE$^\prime$). 
MAE calculates the average absolute difference between actual and predicted values, providing a direct measure of prediction accuracy.
MAPE$^\prime$ modifies 
traditional MAPE to provide a more balanced and equitable error 
metric, particularly effective when actual values are near zero, 
by ensuring symmetry in error treatment and reducing sensitivity 
to extreme values\cite{goodwin1999asymmetry}. MAE and MAPE$^\prime$ are defined as
\begin{gather}
{\rm MAE} = \frac{1}{H}\sum_{j=t}^{t+H-1}\frac{1}{D}\sum_{i=0}^{D-1}|\hat{Y}_j^{(i)}-Y_j^{(i)}|, \\
{\rm MAPE}^\prime = \frac{\sum_{j=t}^{t+H-1}\sum_{i=0}^{D-1}|\hat{Y}_j^{(i)}-Y_j^{(i)}|}{\sum_{j=t}^{t+H-1}\sum_{i=0}^{D-1}|\hat{Y}_j^{(i)}|+|Y_j^{(i)}|},
\end{gather}
where $\hat{Y}_j^{(i)}$ and $Y_j^{(i)}$ are respectively the predicted and actual values at the $j$-th time step and $i$-th dimension variable, $H$ represents the number of forecasting time
steps, and $D$ represents the dimension of variables.
Evidently, lower values of MAE and MAPE$^\prime$ indicate higher forecasting accuracy. Following the same evaluation procedure  
used in the previous studies \cite{chen2023tsmixer,zhang2022crossformer,woo2022etsformer}, we 
compute both metrics on z-score normalized data to measure different variables on the same scale.

\subsection{Baseline Models} \label{sec:III-C}
We demonstrate 
ERKG
by applying it to three state-of-the-art 
prediction models, i.e., Crossformer \cite{zhang2022crossformer},  
ETSformer \cite{woo2022etsformer} and TSMixer \cite{ekambaram2023tsmixer}. They are 
capable of managing complex intra-variable and inter-variable patterns in load 
series.


\paragraph{Crossformer} Crossformer \cite{zhang2022crossformer} 
\qh{utilizes} cross-dimension dependency for multivariates time series forecasting, which embed the input into 2D time and dimension information. 
\qh{We} use the official model code  \footnote{Crossformer: \url{https://github.com/Thinklab-SJTU/Crossformer}}  and hyperparameter settings to train the model.

\paragraph{ETSformer} ETSformer \cite{woo2022etsformer} introduces a novel approach to time series forecasting by leveraging the principles of exponential smoothing,
validating its robustness and efficiency in forecasting tasks. 
We trained the model using the official implementation 
\footnote{ETSformer: \url{https://github.com/salesforce/ETSformer}} 
and hyperparameters.

\paragraph{TSMixer} 
TsMixer \cite{ekambaram2023tsmixer} is a lightweight MLP-Mixer model outperforming complex transformer models with minimal computing usage. 
The model code from HuggingFace  
\footnote{TSMixer: \url{https://huggingface.co/docs/transformers}}. 
\begin{table*}[htbp]
  \centering
  \caption{Comparison of forecasting errors between the baselines and our method}
        \begin{tabular}{cc||cccccccccccc}
    \toprule
    \multicolumn{2}{c}{Method} & \multicolumn{2}{c}{Crossformer} & \multicolumn{2}{c}{ +ours} & \multicolumn{2}{c}{Etsformer} & \multicolumn{2}{c}{ +ours} & \multicolumn{2}{c}{TS-mixer} & \multicolumn{2}{c}{ +ours} \\
    \midrule
    \multicolumn{2}{c}{Metric} & \multicolumn{1}{l}{MAE} & \multicolumn{1}{l}{MAPE$^\prime$} & \multicolumn{1}{l}{MAE} & \multicolumn{1}{l}{MAPE$^\prime$} & \multicolumn{1}{l}{MAE} & \multicolumn{1}{l}{MAPE$^\prime$} & \multicolumn{1}{l}{MAE} & \multicolumn{1}{l}{MAPE$^\prime$} & \multicolumn{1}{l}{MAE} & \multicolumn{1}{l}{MAPE$^\prime$} & \multicolumn{1}{l}{MAE} & \multicolumn{1}{l}{MAPE$^\prime$} \\
    \midrule
    \midrule
    \multicolumn{1}{c|}{\multirow{10}[2]{*}{\begin{sideways}Ampds2\end{sideways}}} & 1     & 0.520  & 1.074  & \textbf{0.352 } & \textbf{0.669 } & 0.443  & 0.902  & \textbf{0.415 } & \textbf{0.826 } & 0.393  & 0.769  & \textbf{0.379 } & \textbf{0.679 } \\
    \multicolumn{1}{c|}{} & 6     & 0.476  & 0.939  & \textbf{0.400 } & \textbf{0.777 } & 0.496  & 1.014  & \textbf{0.445 } & \textbf{0.881 } & 0.425  & 0.885  & \textbf{0.396 } & \textbf{0.738 } \\
    \multicolumn{1}{c|}{} & 12    & 0.491  & 1.038  & \textbf{0.402 } & \textbf{0.776 } & 0.507  & 1.030  & \textbf{0.448 } & \textbf{0.888 } & 0.434  & 0.923  & \textbf{0.402 } & \textbf{0.757 } \\
    \multicolumn{1}{c|}{} & 24    & 0.480  & 0.977  & \textbf{0.407 } & 0.775  & 0.504  & 1.033  & \textbf{0.452 } & \textbf{0.894 } & 0.435  & 0.925  & \textbf{0.404 } & \textbf{0.762 } \\
    \multicolumn{1}{c|}{} & 36    & 0.508  & 1.092  & \textbf{0.417 } & \textbf{0.810 } & 0.504  & 1.037  & \textbf{0.454 } & \textbf{0.890 } & 0.439  & 0.930  & \textbf{0.410 } & \textbf{0.780 } \\
    \multicolumn{1}{c|}{} & 48    & 0.515  & 1.136  & \textbf{0.417 } & \textbf{0.821 } & 0.504  & 1.037  & \textbf{0.455 } & \textbf{0.890 } & 0.441  & 0.930  & \textbf{0.410 } & \textbf{0.776 } \\
    \multicolumn{1}{c|}{} & 60    & 0.491  & 0.991  & \textbf{0.417 } & \textbf{0.778 } & 0.512  & 1.033  & \textbf{0.451 } & \textbf{0.869 } & 0.440  & 0.921  & \textbf{0.411 } & \textbf{0.769 } \\
    \multicolumn{1}{c|}{} & 72    & 0.513  & 1.111  & \textbf{0.421 } & \textbf{0.774 } & 0.511  & 1.039  & \textbf{0.456 } & \textbf{0.893 } & 0.445  & 0.938  & \textbf{0.417 } & \textbf{0.802 } \\
    \multicolumn{1}{c|}{} & 168   & 0.446  & 0.891  & \textbf{0.431 } & \textbf{0.837 } & 0.522  & 1.038  & \textbf{0.465 } & \textbf{0.888 } & 0.455  & 0.957  & \textbf{0.425 } & \textbf{0.798 } \\
    \multicolumn{1}{c|}{} & 336   & 0.472  & 0.954  & \textbf{0.439 } & \textbf{0.772 } & 0.522  & 1.042  & \textbf{0.477 } & \textbf{0.905 } & 0.470  & 0.969  & \textbf{0.438 } & \textbf{0.814 } \\
    \midrule
    \multicolumn{1}{c|}{\multirow{10}[2]{*}{\begin{sideways}UK-Daleh1\end{sideways}}} & 1     & 0.370  & 0.789  & \textbf{0.349 } & \textbf{0.705 } & 0.485  & 0.962  & \textbf{0.464 } & \textbf{0.933 } & 0.362  & \textbf{0.694 } & 0.361  & 0.697  \\
    \multicolumn{1}{c|}{} & 6     & \textbf{0.425 } & \textbf{0.828 } & 0.428  & 0.862  & 0.563  & 1.069  & \textbf{0.536 } & \textbf{1.037 } & 0.446  & 0.868  & \textbf{0.427 } & \textbf{0.817 } \\
    \multicolumn{1}{c|}{} & 12    & 0.464  & 0.937  & \textbf{0.447 } & \textbf{0.911 } & 0.585  & 1.089  & \textbf{0.545 } & \textbf{1.047 } & 0.457  & 0.888  & \textbf{0.435 } & \textbf{0.831 } \\
    \multicolumn{1}{c|}{} & 24    & 0.456  & 0.869  & \textbf{0.450 } & \textbf{0.869 } & 0.596  & 1.110  & \textbf{0.553 } & \textbf{1.069 } & 0.469  & 0.907  & \textbf{0.443 } & \textbf{0.831 } \\
    \multicolumn{1}{c|}{} & 36    & 0.477  & 0.940  & \textbf{0.451 } & \textbf{0.827 } & 0.611  & 1.137  & \textbf{0.562 } & \textbf{1.075 } & 0.485  & 0.938  & \textbf{0.454 } & \textbf{0.855 } \\
    \multicolumn{1}{c|}{} & 48    & 0.476  & 0.979  & \textbf{0.458 } & \textbf{0.907 } & 0.588  & 1.112  & \textbf{0.541 } & \textbf{1.066 } & 0.464  & 0.916  & \textbf{0.443 } & \textbf{0.850 } \\
    \multicolumn{1}{c|}{} & 60    & 0.467  & 0.915  & \textbf{0.460 } & \textbf{0.911 } & 0.592  & 1.126  & \textbf{0.555 } & \textbf{1.079 } & 0.466  & 0.923  & \textbf{0.450 } & \textbf{0.877 } \\
    \multicolumn{1}{c|}{} & 72    & 0.482  & 0.977  & \textbf{0.466 } & \textbf{0.946 } & 0.598  & 1.121  & \textbf{0.548 } & \textbf{1.067 } & 0.470  & 0.931  & \textbf{0.449 } & \textbf{0.872 } \\
    \multicolumn{1}{c|}{} & 168   & 0.493  & 1.078  & \textbf{0.464 } & \textbf{0.975 } & 0.578  & 1.130  & \textbf{0.532 } & \textbf{1.066 } & 0.470  & 0.949  & \textbf{0.446 } & \textbf{0.870 } \\
    \multicolumn{1}{c|}{} & 336   & 0.483  & 1.024  & \textbf{0.457 } & \textbf{0.918 } & 0.571  & 1.137  & \textbf{0.516 } & \textbf{1.057 } & 0.470  & 0.957  & \textbf{0.448 } & \textbf{0.881 } \\
    \midrule
    \multicolumn{1}{c|}{\multirow{10}[2]{*}{\begin{sideways}UmassC\end{sideways}}} & 1     & 0.401  & 0.903  & \textbf{0.384 } & \textbf{0.855 } & 0.495  & 1.047  & \textbf{0.447 } & \textbf{0.934 } & 0.423  & 0.894  & \textbf{0.395 } & \textbf{0.795 } \\
    \multicolumn{1}{c|}{} & 6     & 0.466  & 0.956  & \textbf{0.426 } & \textbf{0.858 } & 0.582  & 1.181  & \textbf{0.513 } & \textbf{1.058 } & 0.484  & 1.047  & \textbf{0.439 } & \textbf{0.922 } \\
    \multicolumn{1}{c|}{} & 12    & 0.482  & 1.001  & \textbf{0.438 } & \textbf{0.900 } & 0.600  & 1.208  & \textbf{0.523 } & \textbf{1.081 } & 0.493  & 1.064  & \textbf{0.441 } & \textbf{0.908 } \\
    \multicolumn{1}{c|}{} & 24    & 0.499  & 1.053  & \textbf{0.468 } & \textbf{0.999 } & 0.598  & 1.193  & \textbf{0.517 } & \textbf{1.079 } & 0.503  & 1.083  & \textbf{0.451 } & \textbf{0.947 } \\
    \multicolumn{1}{c|}{} & 36    & 0.499  & 1.073  & \textbf{0.451 } & \textbf{0.912 } & 0.605  & 1.209  & \textbf{0.534 } & \textbf{1.076 } & 0.505  & 1.093  & \textbf{0.445 } & \textbf{0.901 } \\
    \multicolumn{1}{c|}{} & 48    & 0.493  & 1.047  & \textbf{0.447 } & \textbf{0.899 } & 0.603  & 1.204  & \textbf{0.523 } & \textbf{1.060 } & 0.502  & 1.094  & \textbf{0.452 } & \textbf{0.945 } \\
    \multicolumn{1}{c|}{} & 60    & 0.497  & 1.026  & \textbf{0.457 } & \textbf{0.951 } & 0.616  & 1.214  & \textbf{0.535 } & \textbf{1.092 } & 0.509  & 1.110  & \textbf{0.457 } & \textbf{0.971 } \\
    \multicolumn{1}{c|}{} & 72    & 0.508  & 1.106  & \textbf{0.468 } & \textbf{0.977 } & 0.620  & 1.235  & \textbf{0.544 } & \textbf{1.092 } & 0.509  & 1.109  & \textbf{0.457 } & \textbf{0.955 } \\
    \multicolumn{1}{c|}{} & 168   & 0.501  & 1.068  & \textbf{0.462 } & \textbf{0.940 } & 0.627  & 1.237  & \textbf{0.521 } & \textbf{1.072 } & 0.514  & 1.125  & \textbf{0.456 } & \textbf{0.942 } \\
    \multicolumn{1}{c|}{} & 336   & 0.519  & 1.079  & \textbf{0.487 } & \textbf{1.021 } & 0.623  & 1.219  & \textbf{0.524 } & \textbf{1.059 } & 0.519  & 1.130  & \textbf{0.474 } & \textbf{0.998 } \\
    \midrule
    \multicolumn{1}{c|}{\multirow{10}[2]{*}{\begin{sideways}UmassD\end{sideways}}} & 1     & \textbf{0.263 } & 0.645  & 0.291  & \textbf{0.601 } & 0.339  & 0.722  & \textbf{0.320 } & \textbf{0.678 } & 0.221  & 0.547  & \textbf{0.198 } & \textbf{0.478 } \\
    \multicolumn{1}{c|}{} & 6     & 0.279  & \textbf{0.637 } & \textbf{0.275 } & 0.647  & 0.430  & 0.827  & \textbf{0.415 } & \textbf{0.785 } & 0.265  & 0.646  & \textbf{0.240 } & \textbf{0.574 } \\
    \multicolumn{1}{c|}{} & 12    & 0.311  & 0.745  & \textbf{0.281 } & \textbf{0.668 } & 0.436  & 0.851  & \textbf{0.405 } & \textbf{0.797 } & 0.270  & 0.650  & \textbf{0.255 } & \textbf{0.609 } \\
    \multicolumn{1}{c|}{} & 24    & 0.306  & 0.687  & \textbf{0.280 } & \textbf{0.669 } & 0.483  & 0.881  & \textbf{0.434 } & \textbf{0.810 } & 0.283  & 0.673  & \textbf{0.260 } & \textbf{0.611 } \\
    \multicolumn{1}{c|}{} & 36    & 0.323  & 0.786  & \textbf{0.287 } & \textbf{0.652 } & 0.495  & 0.882  & \textbf{0.452 } & \textbf{0.823 } & 0.292  & 0.707  & \textbf{0.263 } & \textbf{0.625 } \\
    \multicolumn{1}{c|}{} & 48    & 0.328  & 0.774  & \textbf{0.309 } & \textbf{0.677 } & 0.500  & 0.893  & \textbf{0.456 } & \textbf{0.827 } & 0.285  & 0.691  & \textbf{0.261 } & \textbf{0.624 } \\
    \multicolumn{1}{c|}{} & 60    & 0.308  & 0.731  & \textbf{0.302 } & \textbf{0.705 } & 0.500  & 0.893  & \textbf{0.471 } & \textbf{0.838 } & 0.293  & 0.712  & \textbf{0.262 } & \textbf{0.629 } \\
    \multicolumn{1}{c|}{} & 72    & 0.330  & 0.782  & \textbf{0.298 } & \textbf{0.693 } & 0.494  & 0.885  & \textbf{0.464 } & \textbf{0.827 } & 0.297  & 0.711  & \textbf{0.271 } & \textbf{0.640 } \\
    \multicolumn{1}{c|}{} & 168   & 0.364  & 0.867  & \textbf{0.349 } & \textbf{0.731 } & 0.494  & 0.900  & \textbf{0.473 } & \textbf{0.861 } & 0.314  & 0.730  & \textbf{0.291 } & \textbf{0.659 } \\
    \multicolumn{1}{c|}{} & 336   & \textbf{0.360 } & 0.838  & 0.368  & \textbf{0.773 } & 0.484  & 0.889  & \textbf{0.471 } & \textbf{0.853 } & 0.322  & 0.735  & \textbf{0.308 } & \textbf{0.677 } \\
    \midrule
    \multicolumn{2}{c||}{Avg \% Imp} & /     & /     & \textbf{7.85\%} & \textbf{11.67\%} & /     & /     & \textbf{9.12\%} & \textbf{8.81\%} & /     & /     & \textbf{7.27\%} & \textbf{11.49\%} \\
    \bottomrule
    \end{tabular}%
    \label{table:main result}
\end{table*}%

\vspace{-5pt}

\subsection{Experiment Results on Competitive Baselines}\label{main result}
\cxrev{To experimentally demonstrate the effectiveness of the proposed overall knowledge extraction and integration framework, we conducted experiments on competitive baselines, as shown in Section V.C.}
The proposed method is implemented across three baselines with essentially consistent experimental configurations.
\cxreva{For all approaches, the prediction lengths ($H$) are configured from one hour to 336 hours (2 weeks), e.g., $H=336$ denotes predicting the individual load values for each of the future 336 steps.} \cxrev{ Similar to CrossFormer, considering long prediction lengths, we set a constant input length of 336 hours,} \cxrev{i.e., the input $X_{t-L
:t}=[x_{t-L},x_{t-L+1},\dots,x_{t-1}]$ has $L=336$, where $x_{t-1}$  denotes an observation of household and appliances load series at time step $t-1$.} The Adam optimizer is employed with a learning rate of 0.001 and a batch size of 128. Additionally, an early stopping strategy with a patience setting of 10 is incorporated. For other configurations, the original settings specified in their official code repositories are adhered to.
Following the setups in 
\cite{zhang2022crossformer}, a sliding window is utilized with a size equal to the sum of the input length and the prediction length, moving forward step by step, to extract the sample input and the corresponding ground truth from the original datasets. Meanwhile, the same procedure is applied to the state data obtained by Algorithm \ref{clustering algorithm} to acquire the corresponding state  ground truth labels for training the MSP.

\begin{table}[htbp]
  \centering
  \caption{\cxrev{COMPARISON OF FORECASTING ERRORS BETWEEN THE Load Forecasting Methods AND OUR METHOD}}
  \setlength{\tabcolsep}{3pt} 
    \begin{tabular}{@{}ccccccccc@{}} 
    \toprule
    Model & \multicolumn{2}{c}{MTL-GRU} & \multicolumn{2}{c}{+ours} & \multicolumn{2}{c}{VAE-LF} & \multicolumn{2}{c}{+ours} \\
    \midrule
    Metric & mae   & mape  & mae   & mape  & mae   & mape  & mae   & mape \\
    \midrule
    \midrule
    Ampds2 & 0.633  & 1.155  & \textbf{0.592} & \textbf{0.969} & 0.705  & 1.250  & \textbf{0.669} & \textbf{1.020} \\
    UK-Daleh1 & 0.520  & 1.207  & \textbf{0.479} & \textbf{1.078} & 0.517  & 1.180  & \textbf{0.494} & \textbf{1.050} \\
    UmassC & 0.511  & 1.108  & \textbf{0.456} & \textbf{0.896} & 0.542  & 1.230  & \textbf{0.472} & \textbf{0.948} \\
    UmassD & 0.446  & 1.213  & \textbf{0.409} & \textbf{1.100} & 0.461  & 1.468  & \textbf{0.410} & \textbf{1.170} \\
    \midrule
    Avg\%Imp & /     & /     & \textbf{8.31\%} & \textbf{13.72\%} & /     & /     & \textbf{8.22\%} & \textbf{17.20\%} \\
    \bottomrule
    \end{tabular}%
  \label{rlf_compare}%
\end{table}


\textbf{\qh{Results:}} The comparison of forecasting errors between the baselines and the proposed method is shown in Table \ref{table:main result}. 
\cxrevb{MAE and MAPE$^\prime$ exhibit similar patterns.}
Using the MAE metric as an example, the results indicate that in the majority of cases, the proposed method achieves a considerable margin of improvement over these baselines. Crossformer witnesses an average enhancement of 6.14\% and a maximal improvement of 32.22\%; ETSformer experiences an average augmentation of 8.14\% and a peak improvement of 14.77\%;  TSMixer observes an average advancement of 11.83\% and a maximum improvement of 18.05\%. Note that we improved Crossformer MAPE$^\prime$ from 1.074 to 0.669, reducing the error by 37.74\%. \cxrev{Additionally, we adapted load forecasting methods, i.e., MTL-GRU \cite{appliance_analysis_wang2021bottom} and VAE-LF \cite{langevin2023efficient}, by applying ERKG to them for experimentation. The experimental results show that our method reduced the MAE error by an average of 8.31\% (MTL-GRU) and 8.22\% (VAE-LF). The results averaged over multiple prediction lengths are presented in Table \ref{rlf_compare}.}
\subsection{
Visual Analysis of Effectiveness and Robustness in Forecasting}
\begin{figure*}[!t]
\centering
\includegraphics[width=7in]{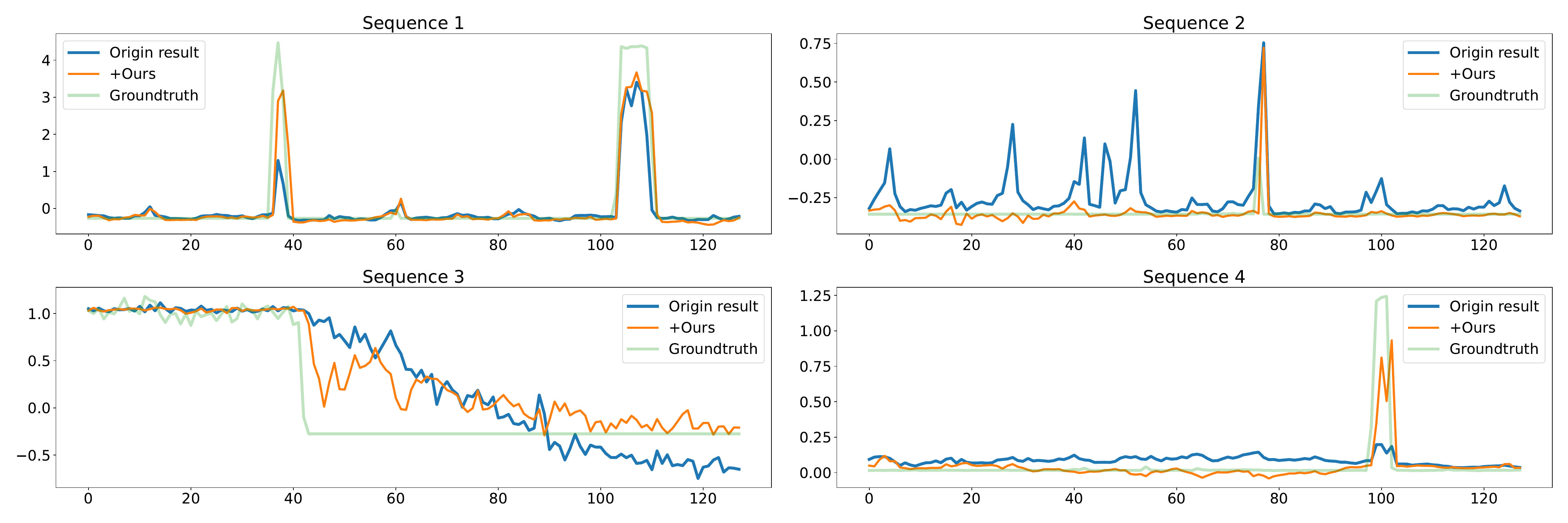}
 \vspace{-10pt}
\caption{Prediction results on UK-Daleh1: We used four prediction sequences to demonstrate the enhanced load forecasting capability of our method under both changing and unchanging \cxrev{appliance operational states.}}
\label{predicted_load_curve}
\end{figure*}

\begin{figure*}[!t]
\centering
\includegraphics[width=6.8in]{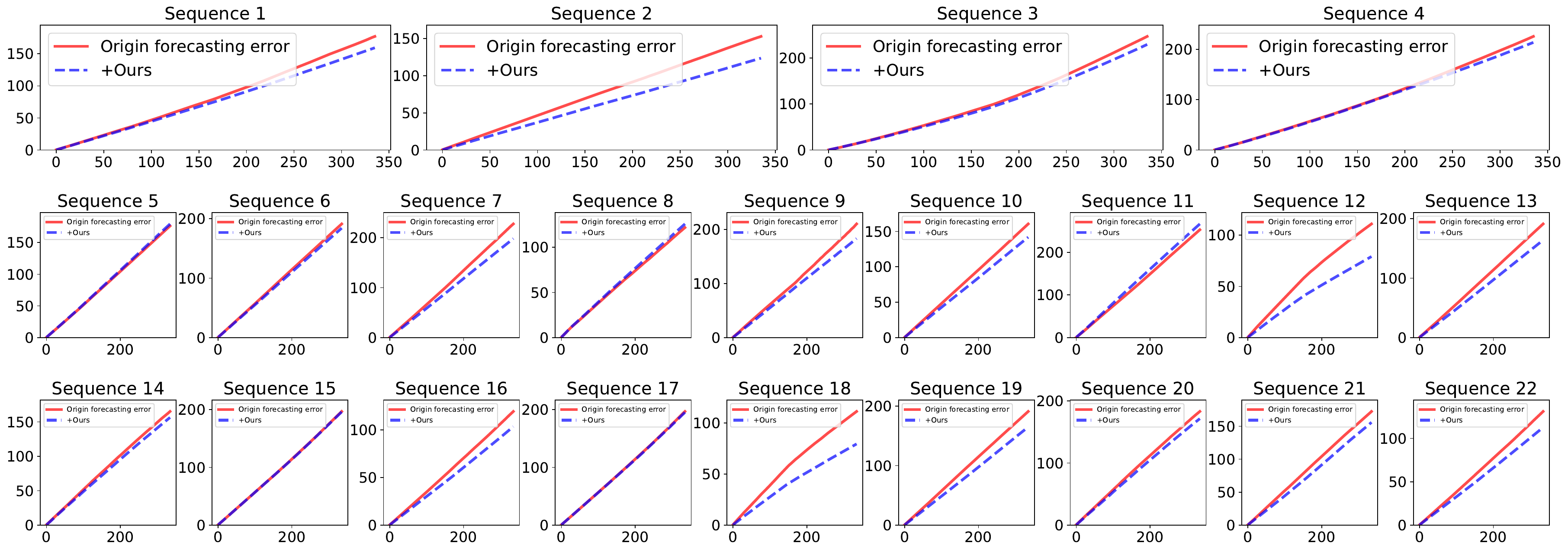}
 \vspace{-10pt}
\caption{Prediction error results on the UMass House D: We visualized the prediction results for all 22 sequences of House D. In these visualizations, lower values indicate smaller errors compared to the actual values. It is evident that after using our method, the prediction errors of the model mostly decreased, with a few remaining stable. Additionally, as the prediction steps increase, the reduction in prediction error becomes more significant.}
\label{cum_forecasting_error}
\end{figure*}

To investigate the effectiveness and robustness of the proposed method, we visualize the refinement of origin predicted load curve with the proposed method (Fig. \ref{predicted_load_curve}), and the change in cumulative prediction error as the forecast step length increases (Fig. \ref{cum_forecasting_error}). In the experiments, the proposed method is implemented with the Crossformer model using the UK-Dale dataset as shown in Fig. \ref{predicted_load_curve}, and with UMass dataset in Fig. \ref{cum_forecasting_error}. The other experimental settings remain the same as those described in Section \ref{main result}. The two graphs are depicted using z-score normalized series data for enhanced clarity and comparison.


\textbf{\qh{Results:}} 
ERKG significantly improves prediction performance. We can observe the effects in Fig. \ref{predicted_load_curve}, which includes both scenarios where events occur (normal patterns) and where no events occur (corruption patterns). Specifically, for scenarios where events occur, i.e., the operational state remains unchanged, such as the 0-60 time steps segment in sequence 2, ERKG improves the forecasting load curve from blue (original) to orange (+ours). This effect is similarly observed in the 0-80 segment of sequence 4. For scenarios where no events occur, i.e., the operational state changes, such as the 30-40 and 100-110 time steps segments in sequence 1, ERKG makes the original predicted peak closer to the actual peak. Similarly effective results can be observed in sequence 3 and sequence 4. These demonstrate that ERKG enables the RLF model to incorporate event knowledge, effectively reducing the impacts from noise.


As depicted in Fig. \ref{cum_forecasting_error}, our analysis at UMass showcases the 
\qh{of the cumulative prediction error.} The horizontal axis represents future prediction steps, while the vertical axis represents the error between model predictions and ground
truth. Here, the red curve depicts the 
error trajectory \qh{of the original model}, 
\qh{in contrast to} the blue curve, which highlights the improvements from the proposed approach. Notably, our approach consistently diminishes prediction errors across various variables at UMass, with a few variables remaining stable. Importantly, error reduction becomes more pronounced over extended time horizons, underscoring the capacity of ERKG to 
boost model robustness.




\subsection{\qh{Ablation Studies}}
\qh{Further,} we perform ablative analysis to explore the 
effects of different components on the performance of \qh{our proposed} ERKG. 
Specifically, 
\qh{ablation studies are conducted to} investigate the effects of the different state predictor and the knowledge guide strategy, respectively.

\textbf{\qh{Effect of State Predictor:}} 
\qh{Two state predictors are experimented}: one using the proposed method with $D$ UEs (+ours), and the other using a state predictor that shares middle layer weights (w/o MSP). The first
group of experiments aims to evaluate the 
efficiency of the proposed MSP in different datasets. 
\qh{We} implement the proposed method by ETSformer (Origin), 
with the prediction length in \qh{$[1, 6, 12, 24, 36, 48, 60, 72, 168, 336]$, where  average performances are shown in}
 Table \ref{tab:blation without MSP}. The 
 results demonstrate that the proposed MSP outperforms the original prediction methods and another state predictor in all comparison results. This confirms the effectiveness of the proposed MSP approach, which involves independently learning the patterns of different load series before understanding the relationships between them. In Table \ref{tab:muti_horizon_without_msp},
 \qh{the} second group of experiments is designed to evaluate the efficiency of the proposed MSP in  different prediction length\qh{s} with the same experiment settings. The experimental results indicate that compared to existing prediction methods and other state predictors, the proposed MSP achieved the best results in all comparison metrics. 
\begin{table}[htbp]
  \centering
  \caption{Ablation without MSP: the average metric which prediction length in \qh{$[1,6,12,24,36,48,60,72,168,336]$.}}
  \setlength{\tabcolsep}{5pt}
    \begin{tabular}{c||cccccc}
    \toprule
    \multicolumn{1}{c}{Method} & \multicolumn{2}{c}{Origin} & \multicolumn{2}{c}{w/o MSP} & \multicolumn{2}{c}{ +ours} \\
    \midrule
    \multicolumn{1}{c}{Metric} & MAE   & MAPE$^\prime$  & MAE   & MAPE$^\prime$  & MAE   & MAPE$^\prime$ \\
    \midrule
    \midrule
    Ampds2 & 0.491  & 1.005  & 0.495  & 0.999  & \textbf{0.452 } & \textbf{0.883 } \\
    UK-Daleh1 & 0.576  & 1.101  & 0.574  & 1.089  & \textbf{0.535 } & \textbf{1.050 } \\
    UMassC & 0.594  & 1.193  & 0.584  & 1.177  & \textbf{0.518 } & \textbf{1.060 } \\
    UMassD & 0.471  & 0.866  & 0.453  & 0.850  & \textbf{0.436 } & \textbf{0.810 } \\
    \bottomrule
    \end{tabular}%
  \label{tab:blation without MSP}%
\end{table}%
\begin{table}[htbp]
  \centering
  \caption{ABLATION WITHOUT MSP: We conduct it on Ampds2 with HORIZON \qh{$[1,6,12,24,36,48,60,72,168,336]$.}}
    \begin{tabular}{c||cccccc}
    \toprule
    \multicolumn{1}{c}{Method} & \multicolumn{2}{c}{Origin} & \multicolumn{2}{c}{w/o MSP} & \multicolumn{2}{c}{ +ours} \\
    \midrule
    \multicolumn{1}{c}{Metric} & MAE   & MAPE$^\prime$  & MAE   & MAPE$^\prime$  & MAE   & MAPE$^\prime$ \\
    \midrule
    \midrule
    1     & 0.440  & 0.884  & 0.452  & 0.906  & \textbf{0.415 } & \textbf{0.826 } \\
    6     & 0.479  & 0.992  & 0.490  & 0.984  & \textbf{0.445 } & \textbf{0.881 } \\
    12    & 0.494  & 1.015  & 0.495  & 1.011  & \textbf{0.448 } & \textbf{0.888 } \\
    24    & 0.490  & 1.019  & 0.496  & 1.004  & \textbf{0.452 } & \textbf{0.894 } \\
    36    & 0.496  & 1.009  & 0.497  & 1.017  & \textbf{0.454 } & \textbf{0.890 } \\
    48    & 0.497  & 1.021  & 0.498  & 1.014  & \textbf{0.455 } & \textbf{0.890 } \\
    60    & 0.495  & 1.018  & 0.499  & 1.018  & \textbf{0.451 } & \textbf{0.869 } \\
    72    & 0.500  & 1.021  & 0.500  & 1.010  & \textbf{0.456 } & \textbf{0.893 } \\
    168   & 0.506  & 1.024  & 0.508  & 1.006  & \textbf{0.465 } & \textbf{0.888 } \\
    336   & 0.514  & 1.049  & 0.516  & 1.023  & \textbf{0.477 } & \textbf{0.905 } \\
    \bottomrule
    \end{tabular}%
  \label{tab:muti_horizon_without_msp}%
\end{table}%

\begin{table}[htbp]
  \centering
  \caption{Ablation without Response Loss: the average metric which horizon in \qh{$[1,6,12,24,36,48,60,72,168,336]$.}}
    \setlength{\tabcolsep}{5pt}
    \begin{tabular}{c||cccccc}
    \toprule
    \multicolumn{1}{c}{Method} & \multicolumn{2}{c}{Origin} & \multicolumn{2}{c}{w/o Response Loss} & \multicolumn{2}{c}{ +ours} \\
    \midrule
    \multicolumn{1}{c}{Metric} & MAE   & MAPE$^\prime$  & MAE   & MAPE$^\prime$  & MAE   & MAPE$^\prime$ \\
    \midrule
    \midrule
    Ampds2 & 0.491  & 1.005  & 0.516  & 1.045  & \textbf{0.452 } & \textbf{0.883 } \\
    UK-Daleh1 & 0.576  & 1.101  & 0.580  & 1.104  & \textbf{0.535 } & \textbf{1.050 } \\
    UmassC & 0.594  & 1.193  & 0.600  & 1.196  & \textbf{0.518 } & \textbf{1.060 } \\
    UmassD & 0.471  & 0.866  & 0.469  & 0.866  & \textbf{0.436 } & \textbf{0.810 } \\
    \bottomrule
    \end{tabular}%
  \label{tab:Ablation without response}%
\end{table}%

\begin{table}[htbp]
  \centering
  \caption{ABLATION WITHOUT Response loss: We conduct it on Ampds2 with HORIZON \qh{$[1,6,12,24,36,48,60,72,168,336]$.}}
    \begin{tabular}{c||cccccc}
    \toprule
    \multicolumn{1}{c}{Method} & \multicolumn{2}{c}{Origin} & \multicolumn{2}{c}{w/o Response Loss} & \multicolumn{2}{c}{ +ours} \\
    \midrule
    \multicolumn{1}{c}{Metric} & MAE   & MAPE$^\prime$  & MAE   & MAPE$^\prime$  & MAE   & MAPE$^\prime$ \\
    \midrule
    \midrule
    1     & 0.440  & 0.884  & 0.491  & 1.041  & \textbf{0.415 } & \textbf{0.826 } \\
    6     & 0.479  & 0.992  & 0.499  & 1.018  & \textbf{0.445 } & \textbf{0.881 } \\
    12    & 0.494  & 1.015  & 0.507  & 1.025  & \textbf{0.448 } & \textbf{0.888 } \\
    24    & 0.490  & 1.019  & 0.502  & 1.019  & \textbf{0.452 } & \textbf{0.894 } \\
    36    & 0.496  & 1.009  & 0.510  & 1.036  & \textbf{0.454 } & \textbf{0.890 } \\
    48    & 0.497  & 1.021  & 0.514  & 1.063  & \textbf{0.455 } & \textbf{0.890 } \\
    60    & 0.495  & 1.018  & 0.517  & 1.067  & \textbf{0.451 } & \textbf{0.869 } \\
    72    & 0.500  & 1.021  & 0.517  & 1.047  & \textbf{0.456 } & \textbf{0.893 } \\
    168   & 0.506  & 1.024  & 0.538  & 1.089  & \textbf{0.465 } & \textbf{0.888 } \\
    336   & 0.514  & 1.049  & 0.564  & 1.043  & \textbf{0.477 } & \textbf{0.905 } \\
    \bottomrule
    \end{tabular}%
  \label{tab:ABLATION WITHOUT Response loss}%
\end{table}%

\textbf{\qh{Effect of Knowledge Guide Strategy:}}
\qh{Our proposed method utilizes a knowledge-guided mechanism based on event-response (+ours). Hence, we design the corresponding ablation study by} comparing with a feature-based knowledge-guided approach (w/o Response Loss), which merges hidden feature maps with the model's hidden feature maps for knowledge guidance \cite{xiao2021face}. The proposed method, implemented via ETSformer (Origin), is evaluated in two groups of experiments as shown in Table \ref{tab:Ablation without response} and \ref{tab:ABLATION WITHOUT Response loss}. The first group assesses the efficiency of the proposed response-based knowledge transfer across different datasets, while the second group examines its efficiency across varying prediction lengths under the same experimental setup. The results show that the proposed method consistently outperforms in all comparison metrics,
\qh{validating} the effectiveness of the proposed knowledge-guided mechanism. 
\cxrev{Additionally, ablation experiments on the number of clusters for electrical appliances
verify the effectiveness of 
the silhouette score to determine the optimal number of operational states, as detailed in Table~\ref{clusterk}.}

\begin{table}[htbp]
  \centering
  \caption{\cxrev{Ablation Study on the Number of Appliance Operational State Clusters ($K$)}}
    \begin{tabular}{c||ccccc}
    \toprule
    \multicolumn{1}{l}{Prediction Length} & K=2 & K=3 & K=4 & K=5 & +ours \\
    \midrule
    1     & 0.487  & 0.479  & 0.466  & 0.462  & \textbf{0.447 } \\
    6     & 0.561  & 0.554  & 0.540  & 0.534  & \textbf{0.513 } \\
    12    & 0.586  & 0.565  & 0.557  & 0.550  & \textbf{0.523 } \\
    24    & 0.589  & 0.569  & 0.557  & 0.549  & \textbf{0.517 } \\
    36    & 0.592  & 0.571  & 0.557  & 0.549  & \textbf{0.534 } \\
    48    & 0.585  & 0.575  & 0.566  & 0.551  & \textbf{0.523 } \\
    60    & 0.600  & 0.582  & 0.577  & 0.558  & \textbf{0.535 } \\
    72    & 0.600  & 0.590  & 0.573  & 0.556  & \textbf{0.544 } \\
    168   & 0.604  & 0.588  & 0.573  & 0.562  & \textbf{0.521 } \\
    336   & 0.611  & 0.597  & 0.572  & 0.565  & \textbf{0.524 } \\
    \bottomrule
    \end{tabular}%
  \label{clusterk}%
\end{table}%

\textbf{{Comparison of Computational Cost:}}
\cxrev{We perform ablation analysis to separately explore the additional time cost
of applying our method and the training
cost of our MSP. Accordingly, we conduct  experiments on ETSformer:
\emph{i)}  in Table \ref{compare_cost1}, we compare
the time per epoch ($t$, in seconds), the number of epochs ($N$) required to complete training, and the total training time ($T$, in seconds) before using our method (Origin) and after using our method (+ours). It can be seen that although our method increases the training time for a single epoch, in some cases, e.g., Ampds2, UK-Daleh1, UMassD, our method reduces the number of epochs needed for the model to converge, which can significantly reduce the overall training cost, sometimes even below the original time, e.g., Ampds2.
\emph{ii) in Table \ref{compare_cost2}, we compare  our Events Forecasting Model (MSP) with the original model (Origin). It is evident that the total training time of the MSP is less than that of the original prediction model in Ampds2, UK-Daleh1 and UMassC.}}

\cxrev{In existing datasets, a significant increase in data size may accelerate model convergence, resulting in training time not rising proportionally, as shown by Ampds2 in Table \ref{compare_cost1}. } \cxrev{There are two potential challenges for the implementation of ERKG in generic real-world setups. Firstly, although ERKG exhibits advantageous performance for individual households, it lacks generalizability across different residents, \textit{e.g.}, a model trained on one household may need retraining when  applied to another due to variations in the quantity and types of appliances. 
\cxrevb{Hence, applying ERKG across different households may increase training costs. Additionally, while ERKG effectively learns event-related associations among appliances, its computational cost rises correspondingly with higher data dimensions.}
In future works, it would be interesting to leverage knowledge \cxrev{distillation}\cite{non_invasive_knowledge_hinton2015distilling} to achieve  light-weighting in modelling an optimization.}


\vspace{-10pt} 
\begin{table}[htbp]
  \centering
 \caption{ \cxrev{COMPARISON OF TRAINING TIME BETWEEN OUR METHOD AND THE ORIGINAL METHOD.}}
    \begin{tabular}{ccccc}
    \toprule
    \multicolumn{1}{l}{Datasets} & Ampds2 &  UK-Daleh1 & UmassC & UmassD \\
    \midrule
    Origin & 577.49  & 757.43  & 122.24  & 165.81  \\
    ($T=N*t$) & (48$*$12.03) & (65$*$11.65) & (23$*$5.31) & (27$*$6.14) \\
    +ours & \textbf{431.80}  & 886.27  & 334.71  & 175.89  \\
    ($T=N*t$) & (\textbf{28}$*$15.42) & (\textbf{56}$*$15.82) & (48$*$6.97) & (\textbf{18}$*$9.77) \\
    \bottomrule
    \end{tabular}%
  \label{compare_cost1}%
\end{table}%

\vspace{-10pt} 

\begin{table}[htbp]
  \centering
\caption{  \cxrev{TRAINING TIME COMPARISON BETWEEN EVENTS FORECASTING AND ORIGINAL METHODS.}}
    \begin{tabular}{ccccc}
    \toprule
    \multicolumn{1}{l}{Datasets} & Ampds2 &  UK-Daleh1 & UmassC & UmassD \\
    \midrule
    Origin & 577.49  & 757.43  & 122.24  & 165.81  \\
    ($T=N*t$) & (48$*$12.03) & (65$*$11.65) & (23$*$5.31) & (27$*$6.14) \\
    MSP & \textbf{198.640 }& \textbf{419.012} & \textbf{119.59 } & 232.07  \\
    ($T=N*t$) & (\textbf{12}$*$16.55) & (\textbf{14}$*$29.93) & (\textbf{19}$*$6.29) & (\textbf{12}$*$19.34) \\
    \bottomrule
    \end{tabular}%
  \label{compare_cost2}%
\end{table}%

\vspace{-10pt} 

\FloatBarrier
 \vspace{-10pt}
\section{Conclusion}\label{sec:section6}
In this work, we propose ERKG, a knowledge guided approach
to enhance RLF models for both household-level
and application-level electricity usage predictions. 
\qh{In  ERKG, it}  learns event-related sparse knowledge from dense load series
\qh{by leveraging a novel forecasting model for estimating electricity usage events 
and a corresponding} knowledge-guided mechanism based on event response. \qh{With sparse knowledge from event response, ERKG} 
are more effective at learning electricity usage event than the existing multivariates time series forecasting modle. 
Notably, ERKG can serve as a plug-in block during the forecasting model training stage, allowing existing models to focus more on sparse event information 
but less  on noise, thereby improving performances. Extensive numerical experiments demonstrate that ERKG effectively enhances the RLF performance of existing state-of-the-art models, with an average improvement of 9\% in MAE. Future directions could include considering covariates that influence events, such as  exploring unlabeled modeling of the relationships between events, \cxrev{and considering how to ensure that the model’s computational cost, and the number of parameters remains within an \cxrev{appropriate} range as the data size increases.}

\section*{Acknowledgments}
The work of Xin Cao and Yingjie Zhou was supported in part by National Natural Science Foundation of China (NSFC) under Grant No. 62171302, the 111 Project under Grant No. B21044 and Sichuan Science and Technology Program under Grant No. 2023NSFSC1965.

\bibliographystyle{IEEEtran}
\bibliography{ref}

\end{document}